\pdfoutput=1

\documentclass[11pt]{article}

\usepackage[preprint]{acl}

\usepackage{times}
\usepackage{latexsym}

\usepackage[T1]{fontenc}

\usepackage[utf8]{inputenc}

\usepackage{microtype}

\usepackage{inconsolata}

\usepackage{graphicx}
\usepackage{enumitem}
\usepackage{fancyvrb}
\usepackage{listings}
\usepackage{multirow}
\usepackage{multicol}
\usepackage{booktabs}
\usepackage{amsmath}
\usepackage{amsfonts}
\usepackage{todonotes}

\usepackage[most]{tcolorbox}
\usepackage{xcolor}
\usepackage[dvipsnames]{xcolor}
\usepackage{arydshln}

\definecolor{c1}{cmyk}{0,0.6175,0.8848,0.1490} 
\definecolor{c2}{cmyk}{0.1127,0.6690,0,0.4431} 
\definecolor{c3}{cmyk}{0.3081,0,0.7209,0.3255} 
\definecolor{c4}{cmyk}{0.6765,0.2017,0,0.0667} 
\definecolor{c5}{cmyk}{0,0.8765,0.7099,0.3647} 
\definecolor{forestgreen}{HTML}{397727} 

\usepackage{url}

\newtcbox{\hlprimarytab}{on line, rounded corners, box align=base, 
  colback=c3!10, colframe=white, size=fbox, arc=3pt, 
  before upper=\strut, top=-1pt, bottom=-4pt, left=-2pt, right=-2pt, boxrule=0pt}
\newtcbox{\hlsecondarytab}{on line, box align=base, 
  colback=c1!10, colframe=white, size=fbox, arc=3pt, 
  before upper=\strut, top=-2pt, bottom=-4pt, left=-2pt, right=-2pt, boxrule=0pt}

\title{Repetition over Diversity: High-Signal Data Filtering for Sample-Efficient German Language Modeling}

\author{Ansar Aynetdinov \And Patrick Haller \\
\\
Humboldt-Universität zu Berlin
\\
\texttt{\{aynetdia{\normalfont,} patrick.haller.1{\normalfont,} alan.akbik\}@hu-berlin.de} \And
Alan Akbik \\
}

\begin{document}
\maketitle

\begin{abstract}

Recent research has shown that filtering massive English web corpora into high-quality subsets significantly improves training efficiency. However, for high-resource non-English languages like German, French, or Japanese, aggressive filtering creates a strategic dilemma: should practitioners prioritize \textit{diversity} by training once on large amounts of lightly filtered web data, or prioritize \textit{quality} by strictly filtering for a high-quality core and repeating it over multiple epochs? \\
We investigate this trade-off for German by applying hierarchical quality filters to 500M web documents, comparing multi-epoch training on filtered subsets against single-pass training on a diverse corpus. Our experiments across multiple model scales and token budgets show that repeating high-quality data consistently outperforms single-pass training on larger, less filtered sets. Notably, the performance gap persists even after 7 epochs. Our findings suggest that for non-English LLMs, semantic concentration through quality filtering offers a more viable path to efficient language modeling than simply maximizing unique data volume. \\
We release our German language models
(called \href{https://huggingface.co/Boldt}{\textbf{\textsc{Boldt}}}),
as well as our cleaned evaluation benchmarks to the research community. Our experiments indicate that they achieve state-of-the-art results despite training on 10-360x fewer tokens than comparable models. 

\end{abstract}

\section{Introduction} \label{sec:intro}

The prevailing scaling laws for large language models (LLMs) emphasize a simple triad: more parameters, more compute, and more data \cite{kaplan2020, chinchilla}. However, as the field moves towards more sample-efficient training, the "more is better" paradigm has been increasingly challenged by "quality-first" approaches. Previous works demonstrated that training LLMs on web corpora filtered for high-quality content can yield significantly stronger performance than training on massive, unfiltered datasets \cite{glam2022, phi1, fineweb-edu}.

While this shift is well-documented for English, it presents a unique challenge for non-English high-resource languages such as German, French, Japanese, and Chinese. These languages possess substantial web corpora containing hundreds of billions of tokens, but they lack the multi-trillion-token abundance of English. In these data-constrained scenarios, strict filtering creates a strategic dilemma: should practitioners prioritize diversity by applying only light filters to maintain large token pools (e.g. 100B tokens), or should \emph{semantic density}, i.e. expected training signal per token, be prioritized by applying strict quality filters to obtain a smaller subset (e.g. 25B tokens) and repeating it over multiple epochs?

\noindent \textbf{This Study.} We investigate this trade-off using German as a representative case study for non-English high-resource languages. 
We filter raw web corpora through three hierarchical qualitative tiers:
\textit{Coherence}, which removes structural noise to ensure syntactic flow; \textit{Information Value}, which retains only content-rich and factual documents; and \textit{Educational Quality}, which selects for pedagogical clarity and explanatory depth. We further derive a \textit{Dense-Core} subset -- the intersection of these criteria -- to represent the upper bound of semantic density available in German web data.

By training language models from scratch on these filtered subsets, we isolate the impact of each semantic property on downstream performance and test whether the cautious approach to multi-epoch training prevalent in the literature \cite{muennighoff-2023-repetition,luukkonen-etal-2023-fingpt,croissant-llm} remains warranted when data selection is optimized for knowledge density. We also explore training curricula that transition from diverse data pools to high-density subsets, testing whether data quality staging provides benefits over static data mixtures.

\noindent \textbf{Contributions.} This work makes the following contributions: 
\begin{itemize}[leftmargin=\parindent] 

\item We demonstrate that repeated training on high-quality subsets outperforms maximizing unique web document coverage under a fixed pre-training budget in German language modeling. We specifically analyze the distinct impacts of our three semantic tiers: \textit{Coherence}, \textit{Information Value}, and \textit{Educational Quality}. 

\item We show that multi-epoch pre-training on filtered data positively impacts instruction-tuning behavior, leading to higher correctness in assistant tasks.

\item We identify and correct significant noise in existing German translations of ARC-Challenge, Hellaswag, Lambada, and OpenBookQA benchmarks. We release these updated benchmarks to ensure more reliable assessment of LLM performance in the German NLP community.

\item We release a series of German Small Language Models (SLMs) up to 1B parameters.
Despite being trained on an order of magnitude fewer tokens than competitive baselines, our models achieve state-of-the-art results for their size, underscoring the importance of high-quality data selection in resource-constrained settings.

\end{itemize}

\section{Related Work}

\noindent \textbf{Data Filtering for LLMs.} 
Since GPT-2 and T5 \cite{gpt2, t5}, filtered CommonCrawl \cite{commoncrawl} has been central to LLM pre-training. Datasets like C4, OSCAR, and RefinedWeb employed deduplication, language filtering, and heuristics to remove noisy data \cite{c4, oscar, refined_web}. GPT-3 and Gopher further demonstrated that classifier-based quality filtering significantly improves outcomes \cite{gpt3, gopher}. 

Recent work has further refined model-based filtering: the Phi models \cite{phi1, phi1_5} showed that "textbook-quality" data dramatically improves sample efficiency, while \citet{penedo2024fineweb} further scaled scaled this with the creation of FineWeb and its more restrictive subset, FineWeb-Edu \cite{fineweb-edu}, which uses a classifier to identify documents with high educational value. \citet{peng2025dataman} further refined this by deriving 14 quality criteria and combining them with domain-aware mixing to optimize data selection across heterogeneous source types. Unfortunately, these techniques remain less explored for non-English languages where strict filtering may critically reduce the available token budget.

\begin{table*}[t]
\small
\centering
\resizebox{1\linewidth}{!}{
\begin{tabular}{l ccccccc}
\toprule
\textbf{Subset} & \textbf{N Docs} & \textbf{Yield} & \textbf{Token Count} & \textbf{Doc Length} & \multicolumn{3}{c}{\textbf{Tokenizer Fertility}} \\
 & (Millions) & (\%) & (Tokens) & ($\mu \pm \sigma$) & Train & Test & Benchmarks \\
\midrule
FW2-DE (Full Pool) & 496.0 & 100.0 & - & - & - & - & - \\
\midrule
\textit{Hierarchical Tiers} & & & & & \\
\textsc{Random} (Baseline) & 128.9* & 26.0* & 100B* & 786 $\pm$ 1725 & 1.49 & 1.48 & 1.57 \\
\textsc{Coherence} & 300.6 (138.3*) & 60.6 (27.8*) & 100B* & 730 $\pm$ 1540 & 1.48 & 1.50 & 1.56 \\
\textsc{Information Value} & 43.5 & 8.8 & 65B & 1494 $\pm$ 2561 & 1.36 & 1.42 & 1.42 \\
\textsc{Educational Quality} & 30.2 & 6.1 & 33B & 1087 $\pm$ 2103 & 1.33 & \textbf{1.40} & \textbf{1.38} \\
\midrule
\textit{Target Core} & & & & & \\
\textsc{Dense-Core} (Intersection) & \textbf{24.5} & \textbf{5.1} & \textbf{28B} & \textbf{1150 $\pm$ 2193} & \textbf{1.32} & \textbf{1.40} & \textbf{1.38} \\
\midrule
\textit{External Baselines} & & & & & \\
FW2-MKC \cite{messmer2025enhancing} & 43.2 & 8.7 & 35B & 823 $\pm$ 1895 & 1.33 & \textbf{1.40} & \textbf{1.38} \\
AA High \cite{burns2025-aa-germanweb} & 70.4 & 14.2 & 21B & 296 $\pm$ 368 & 1.35 & \textbf{1.40} & 1.43 \\
\bottomrule
\end{tabular}
}
\caption{Dataset statistics for the German split of FineWeb-2 (FW2-DE) and derived subsets. \textbf{Yield} represents the percentage of documents retained from the original pool. \textbf{Doc Length} is expressed in tokens. \textsc{Coherence} reports document statistics for both the full subset and the 100B sample (denoted with *) used in further experiments.}
\vspace{-4mm}
\label{tab:filter_statistics}
\end{table*}

\noindent \textbf{Overcoming the data bottleneck.} 
As high-quality data demand outpaced the growth of available corpora, practitioners incorporated code, social media, books, and scientific papers into pre-training mixtures \cite{pile, dolma}, and explored synthetic data generation for pre-training data enrichment \cite{kang-etal-2025-demystifying}, despite the dangers of model collapse \cite{model_collapse}.

\citet{muennighoff-2023-repetition} investigated scaling laws in data-constrained regimes, concluding that the returns from repeated exposure to the same dataset start to diminish after 4 epochs. \citet{fang2026datasets} extend this by showing that repeatedly training on an aggressively filtered dataset for up to 10 epochs can outperform a single pass over a 10x larger unfiltered superset. Our work builds upon these findings by investigating whether strict quality filtering presents a practical obstacle in non-English settings where the unfiltered data pool is substantially smaller, using German as a case study.

\noindent \textbf{Non-English Language Modeling.} 
While English dominates the LLM landscape, several efforts have targeted monolingual or highly specialized multilingual models for high-resource languages like Japanese and German \cite{llm-jp, llamlein}. A lot of the progress in non-English language modeling, however, comes from either continued-pre-training of English-centric models \cite{zheng-etal-2024-cross-lingual, llama-genba, kuulmets-etal-2024-teaching} or pre-training on massive multilingual dataset mixtures \cite{bloom, gemma3, llama3, qwen3}.

In the context of German, two recent projects have proposed specialized filtering strategies for the German subset of the multilingual FineWeb2 dataset \cite{penedo2025fineweb2}. \citet{burns2025-aa-germanweb} developed Aleph-Alpha-GermanWeb (AA-High), using a multi-dimensional classifier to bucket documents by grammar and style. Separately, \citet{messmer2025enhancing} introduced FW2-MKC, which uses an embedding-based approach to select web documents that align with established knowledge benchmarks and instruction-tuning sets. In our experiments, we use these two datasets as external benchmarks to compare against our own hierarchical qualitative filters.

\section{Data and Filtering} \label{sec:data}

Our data pipeline is built upon the German split of the FineWeb2 (FW2-DE) dataset \cite{penedo2025fineweb2}. While FW2-DE provides a foundation of deduplicated web content, its quality-based filtering is rather basic. To enable our study of semantic density, we define a hierarchical filtering framework that moves from structural surface features to deep pedagogical value. We deliberately scope our filtering to three interpretable tiers rather than a larger set of fine-grained criteria, as our goal is not to propose a novel filtering algorithm, but to investigate the quality-quantity trade-offs imposed by non-English data constraints.

\subsection{Hierarchical Qualitative Filters} \label{sec:qual_filters}

Motivated by FineWeb-Edu's \cite{fineweb-edu} success in English, we manually inspected Fineweb-Edu samples across Educational Quality tiers and found that while most documents scoring low on this quality metric did suffer from structural or informational shortcomings, a non-negligible amount contained useful information despite lacking pedagogical framing. This observation led us to define a hierarchy of filters, moving from coarse to strict, in order to isolate and capture these different levels of filtering strictness individually rather than conflating them into a single criterion.

We implement three document-level classifiers to score the FW2-DE pool. These filters are not mutually exclusive but represent increasing tiers of selection strictness. Details on the classifier training pipeline are provided in Appendix~\ref{app:filter_details}.

\noindent 
\textbf{\textsc{Coherence}.} This filter targets basic linguistic and structural integrity. It is designed to remove "word-salad" documents, truncated HTML exports, and fragmentary snippets. A high coherence score indicates a document with natural syntactic flow, regardless of whether the content is informative.

\noindent 
\textbf{\textsc{Information Value}.} This tier selects for "signal density." It identifies documents that are fact-bearing and content-rich, like e.g. technical reports, news articles, or specialized documentation, while filtering out generic web prose, SEO-heavy landing pages, and repetitive boilerplate.

\noindent 
\textbf{\textsc{Educational Quality}.} This is our most restrictive tier, modeled after the criteria used in FineWeb-Edu. It prioritizes textbook-like clarity and pedagogical value, selecting for documents that explain concepts or provide structured knowledge suitable for a curriculum.

\noindent 
\textbf{\textsc{Dense-Core}.} By taking the intersection of these three filters, we derive \textsc{Dense-Core}. This subset contains documents that are simultaneously coherent, information-dense, and educational. As shown in Table~\ref{tab:filter_statistics}, this subset represents the most semantically concentrated version of the German web, which we use as our primary high-quality core for multi-epoch training.

\subsection{Dataset Statistics and Tokenization}
 
To ensure fair comparison across data regimes, we train dedicated BPE tokenizers (vocabulary size 32,000) for each subset, optimizing for their specific lexical distributions.

\noindent 
\textbf{Analysis.} 
Table~\ref{tab:filter_statistics} shows that while the \textsc{Coherence} filter is relatively permissive and retains the majority of the documents from the original FW2-DE pool, the \textsc{Educational Quality} filter is substantially more selective. As selection strictness increases, average document length rises markedly -- the \textsc{Dense-Core} subset contains documents that are on average roughly 50\% longer than in the \textsc{Random} subset, which is a 100B token random sample from the full FW2-DE pool, reflecting substantive long-form content rather than fragmented web snippets. 

As for the tokenizer efficiency, we see a clear trend indicating that discarding lower-quality data from the tokenizer training data improves the resulting tokenizer fertility both on the respective train splits, as well as the shared holdout FW2-DE test set. We also report average tokenizer fertility on question prompts from the benchmarks in the evaluation suite used in our experiments, and observe the same trend there.

\begin{figure}[t]
\centering
\begin{tcolorbox}[colback=white,colframe=black]
\textbf{EN (Original)}: [...] Both Olaf and the boy heard it and walked towards \framebox{\textcolor{ForestGreen}{Bob}}

\vspace{1mm}

\textbf{DE (Old)}: [...] Sowohl Olaf als auch der Junge hörten es und \textcolor{red}{gingen} auf Bob \framebox{\textcolor{red}{zu}}

\vspace{1mm}

\textbf{DE (Fixed)}: [...] Sowohl Olaf als auch der Junge hörten es und gingen auf \framebox{\textcolor{ForestGreen}{Bob}} \textcolor{lightgray}{zu}

\end{tcolorbox}
\vspace{-2mm}
\caption{Example of a problematic instance from the German translation of the LAMBADA benchmark in EleutherAI's Evaluation Harness \cite{eval-harness}. We highlight \framebox{\textcolor{ForestGreen}{correct}} and \framebox{\textcolor{red}{incorrect}} target labels.}
\vspace{-4mm}
\label{fig:lambada-problem}
\end{figure}

\section{Evaluation Datasets} \label{sec:benchmarks}

A significant challenge in assessing LLMs in non-English languages is the lack of broad-range evaluation benchmarks. While many seminal benchmarks were devised in English by human experts, their non-English counterparts are often simply machine-translated without any attention to language-specific rules, often produced by older, less capable models like e.g. early versions of GPT-4 \cite{okapi, stablelm_mt}. 
As a result, these versions often contain non-idiomatic phrasing, grammatical errors, and, more critically, structural artifacts that fundamentally alter the difficulty of the original task. As a part of this work, we contribute a modernized and cleaned suite of German benchmarks designed to provide a more reliable signal for model performance.

\subsection{Addressing Translation Artifacts}

In our analysis of existing German benchmark variants, we identified the main issue with completion-based tasks, often used for evaluation of pre-trained LLMs: discrepancies in the \textit{word order} between English and German. 

In completion tasks like HellaSwag and LAMBADA, German's verb-final structures and split verbs frequently move the original "completion target" away from sentence-final position, disrupting the intended prediction task. Refer to Figure~\ref{fig:lambada-problem} for an example of this issue in the LAMBADA benchmark: instead of tracking the entity "Bob" across sentences, the task becomes predicting the prefix of a verb that is two words away.

To address this issue, we re-translated the suite using the state-of-the-art machine translation \textbf{Tower+ 72B} model \cite{rei2025towerplus}. Instead of translating each instance component (e.g. context, target, options) in isolation, which broke German sentence coherence in e.g. a number of HellaSwag samples, we translate full instances jointly and ensure the preservation of the originally intended completion target using automated checks. We discarded the rare instances (<0.5\%), in which machine translation failed or did not preserve the intended task logic due to word order discrepancies between English and German. Table~\ref{tab:mt_benchmarks_stats} summarizes our modernized evaluation suite.

\begin{table}[t]
\centering
\small
\begin{tabular}{lrrr}
\toprule
\textbf{Benchmark} & \textbf{Original} & \textbf{Removed} & \textbf{Modernized} \\
\midrule
Global MMLU & 14,042 & 0 & 0 \\
ARC-Challenge & 1,168 & 0 & 1,168 \\
ARC-Easy & 2,376 & 6 & 2,370 \\
OpenBookQA & 500 & 5 & 495 \\
HellaSwag & 10,042 & 47 & 9,995 \\
LAMBADA & 5,153 & 3 & 5,150 \\
\bottomrule
\end{tabular}
\caption{Statistics for our German evaluation suite. All benchmarks except Global MMLU were re-translated from their original English version. Removed items indicate instances where translation itself failed or broke the task integrity.}
\vspace{-4mm}
\label{tab:mt_benchmarks_stats}
\end{table}

\begin{table*}[ht]
\centering
\resizebox{1\linewidth}{!}{
\begin{tabular}{lcccccccc}
\toprule
\toprule
Subset & Tokens & MMLU & ARC-C & ARC-E & H-Swag & LAMBADA & OBQA & Avg. \\
\midrule
\multicolumn{9}{l}{\small \emph{Baseline}} \\
\textsc{Random} & 100B (1.0x) & 27.13 & 26.15 & 41.10 & 37.18 & 33.52 & 41.01 & 34.35 \\
\hdashline
\multicolumn{9}{l}{\small \emph{Single Filters}} \\
\textsc{Coherence} & 100B (1.0x) & 27.06 & 27.65 & 42.74 & 40.45 & 38.33 & 42.63 & 36.48 \\
\textsc{Information Value} & 65B (1.5x) & 28.29 & 30.46 & 46.20 & 40.71 & 38.52 & 44.04 & 38.04 \\
\textsc{Educational Quality} & 33B (3.0x) & 28.64 & \textbf{31.49} & \textbf{50.91} & 40.57 & 36.49 & 43.64 & 38.62 \\
\hdashline
\multicolumn{9}{l}{\small \emph{Filter Combinations}} \\
\textsc{Dense-Core} & 28B (3.6x) & \textbf{28.97} & 31.40 & 50.55 & \textbf{41.10} & 37.55 & \textbf{45.86} & \textbf{39.24} \\
MKC \cite{messmer2025enhancing} & 35B (2.9x) & 28.00 & 27.37 & 46.37 & 39.49 & \textbf{40.43} & 42.02 & 37.28 \\
\textsc{AA High} \cite{burns2025-aa-germanweb} & 21B (4.8x) & 26.55 & 25.31 & 39.83 & 37.39 & 29.75 & 40.00 & 33.14 \\
\midrule
\multicolumn{9}{l}{\small \emph{Curriculum-Based (100B Budget)}} \\
\textsc{Sorted} & 100B (1.0x) & 28.00 & 30.18 & 46.25 & 40.25 & 35.98 & 42.63 & 37.22 \\
\textsc{Phased} & 78B (1.3x) & 28.52 & 29.34 & 47.89 & 40.35 & 36.06 & 43.67 & 37.64 \\
\midrule
\midrule
\multicolumn{9}{l}{\small \emph{Reference Models (Total Tokens Trained)}} \\
LLäMmlein-120M \cite{llamlein} & 1T & 26.41 & 24.74 & 37.13 & 23.02 & 26.97 & 43.03 & 33.33 \\
Gemma-3-270M \cite{gemma3} & 6T${^*}$ & 26.09 & 24.93 & 34.68 & 31.60 & 29.85 & 37.37 & 32.07 \\
Qwen-3-0.6B-Base \cite{qwen3} & 36T${^*}$ & 29.87 & 32.90 & 41.90 & 38.13 & 39.57 & 41.01 & 37.23 \\
\bottomrule
\bottomrule
\end{tabular}
}
\caption{Benchmark results for 350M models. Tokens refers to unique tokens in the subset, while the bracketed value indicates the number of epochs to reach the 100B budget. Token counts of multilingual data mixtures are denoted with ${^*}$.
}
\vspace{-4mm}
\label{tab:main-results}
\end{table*}

\subsection{Final Evaluation Suite}

The resulting evaluation suite provides a holistic perspective on model performance across factual knowledge, commonsense reasoning, and linguistic context tracking. Factual knowledge is assessed via the German subset of Global MMLU \cite{global_mmlu}, which spans 57 subjects across STEM, the social sciences, and the humanities. Reasoning capabilities are measured using ARC-Easy, ARC-Challenge \cite{arc_challenge}, and OpenBookQA \cite{openbookqa}, which require models to perform grade-school level science inference. Finally, we evaluate commonsense narrative continuation and discourse-level context tracking through HellaSwag \cite{hellaswag} and the OpenAI version of LAMBADA \cite{lambada}. For all evaluations in this paper, we rely on length-normalized conditional log-likelihoods of continuation options to choose the correct answer in multiple choice benchmarks.

Table~\ref{tab:mt_benchmarks_stats} summarizes the composition of the modernized suite and the impact of our manual cleaning process. Notably, the exclusions required to maintain task integrity were minimal, with the largest adjustment occurring in HellaSwag due to its larger size. We release these cleaned benchmarks to the community to facilitate more rigorous and standardized assessment in German NLP.

\section{Investigating the Quality-Quantity Trade-off}

\label{sec:results} In this section, we present a series of experiments designed to isolate the impact of semantic density on model performance. We investigate whether moving away from the traditional "single-pass" pre-training paradigm to repetition of high-quality subsets can compensate for a lack of unique token diversity. 

\subsection{Experiment I: Token Allocation Strategies (100B Budget)} 
The primary objective of this experiment is to determine the optimal way to distribute a fixed training budget of 100B tokens when faced with a choice between broad, unique web data and narrow, high-density educational content.

\subsubsection{Experimental Setup}

We train decoder-only transformers with 350M (non-embedding) parameters following the Llama architecture \cite{llama2}, all restricted to 100B tokens total exposure to ensure a fair comparison.
Details on hyperparameter and optimizer choice are provided in Appendix~\ref{app:hparams}.

To test the interaction between diversity and density, we define three training strategies:

\begin{enumerate}
\item \textbf{Uniform Baselines:} Single-pass training on unique documents from the \textsc{Random} or \textsc{Coherence} pools. For both baselines we randomly sample documents until we reach the 100B limit from the full FW2-DE and \textsc{Coherence} pools of documents respectively.

\item \textbf{Dense Repetition:} Multi-epoch training on the remaining, higher-quality subsets. For instance, for the 28B-token \textsc{Dense Core} core, the 100B budget results in approximately 3.6 training epochs. We also consider the quality-filtered MKC and AA High subsets of FW2-DE devised independently from this work by \citet{messmer2025enhancing} and \citet{burns2025-aa-germanweb} respectively.
\vspace{-2mm}

\item \textbf{Hybrid Curricula:} "Coarse-to-fine" schedules designed to test whether finishing pre-training on higher-quality data can leverage both diversity and quality. The motivation for these curricula stems from the intuition that models might benefit from broad initial exposure to diverse linguistic patterns available in lower-quality data before refining on concentrated knowledge. We implement two variants: a \textsc{Phased} 50/50 split (50B tokens of \textsc{Random} followed by 50B tokens of \textsc{Dense Core}) and a \textsc{Sorted} schedule, where 100B \textsc{Coherence} tokens are presented in ascending order of their Educational score. 

\end{enumerate}

\subsubsection{Results and Analysis}

Table~\ref{tab:main-results} demonstrates significant gains from prioritizing semantic density over document diversity. \textsc{Dense-Core} outperforms the \textsc{Random} baseline by 4.89 points on average while training for approximately 3.6 epochs over 28B unique tokens, suggesting that for information-dense German text, high signal-to-noise ratio of the curated content outweigh potential multi-epoch repetition risks.

\begin{figure}[t]
  \includegraphics[width=\linewidth]{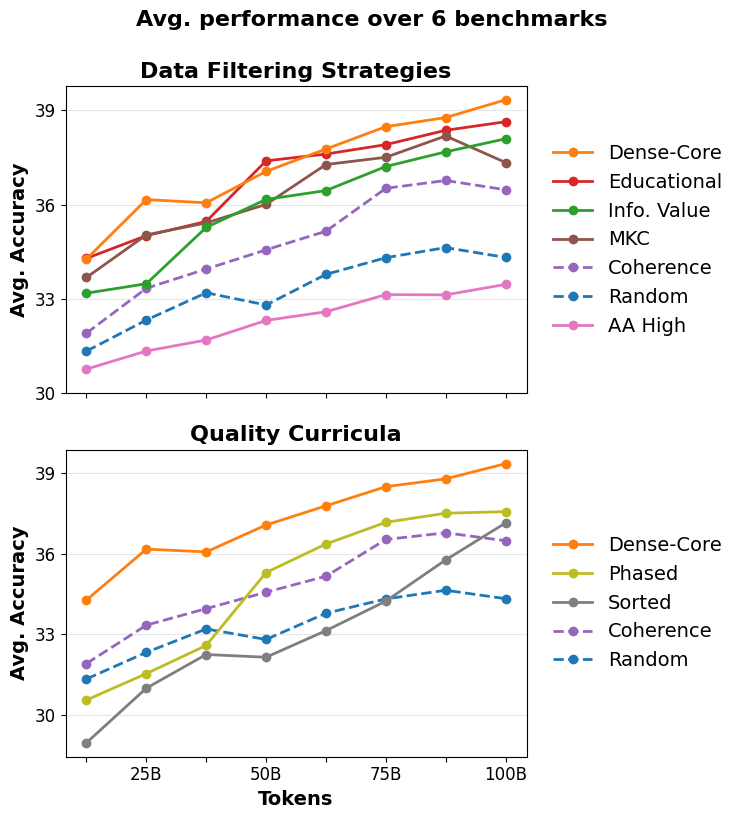}
  \caption{Zero-shot evaluation of 350M models trained on 100B tokens of different FW2-DE subsets over pre-training checkpoints.}
  \vspace{-4mm}
  \label{fig:checkpoints-100B}
\end{figure}

\noindent\textbf{Training dynamics and checkpoint analysis.} 
Figure~\ref{fig:checkpoints-100B} shows that \textsc{Dense-Core} maintains a consistent advantage throughout the entire 100B token trajectory rather than only at convergence. This indicates that the benefit of high-quality repetition is not confined to the final stages of training, but rather provides a steeper learning curve from the onset. The poor performance of the \textsc{AA~High} subset (repeated $4.8\times$) indicates that multi-epoch training only benefits from sufficiently high-quality data.
We attribute its poor downstream impact to usage of smaller and by proxy weaker models as annotators and classifiers for its filtering.

The \textsc{Phased} and \textsc{Sorted} curricula, on the other hand, show a clear inflection point: their performance improves markedly in the second half of training, coinciding with a transition to higher-quality data. While these curricula eventually reach high scores, they consistently trail the pure \textsc{Dense-Core} trajectory, suggesting that even an initial "warm-up" phase of low-signal tokens dilutes the overall information density of a fixed budget.

\noindent \textbf{Comparison to state-of-the-art models.} We compare our 350M models to reference models of a similar but not identical size due to limited availability of models capable of German. We note that our \textsc{Dense-Core} model outperforms the monolinugal \textsc{LLäMmlein-120M} and multilingual \textsc{Gemma-3-270M} and \textsc{Qwen-3-0.6B-Base} \cite{llamlein, gemma3, qwen3} on our evaluation suite, despite these models being trained on $10\times$, $60\times$, and $360\times$ more tokens in total respectively, with the latter two models including German only as a fraction of the total token budget. This suggests that targeted high-signal dataset cores enable performance typically requiring far larger compute and data budgets.

\subsection{Experiment II: Parameter Scaling and Efficiency}

To verify that the performance gains observed in the 350M regime generalize as model capacity increases, we scale our primary experiments to 1B parameter models (see Appendix \ref{app:hparams} for details).

\subsubsection{Experimental Setup}

We focus on the two most distinct strategies: the \textsc{Random} baseline representing high-coverage, single-pass training, and the \textsc{Dense-Core} representing high-density repetition. Both models are trained on the same 100B token budget as in Experiment I.

\subsubsection{Results and Analysis}

\begin{figure}[t]
  \includegraphics[width=\linewidth]{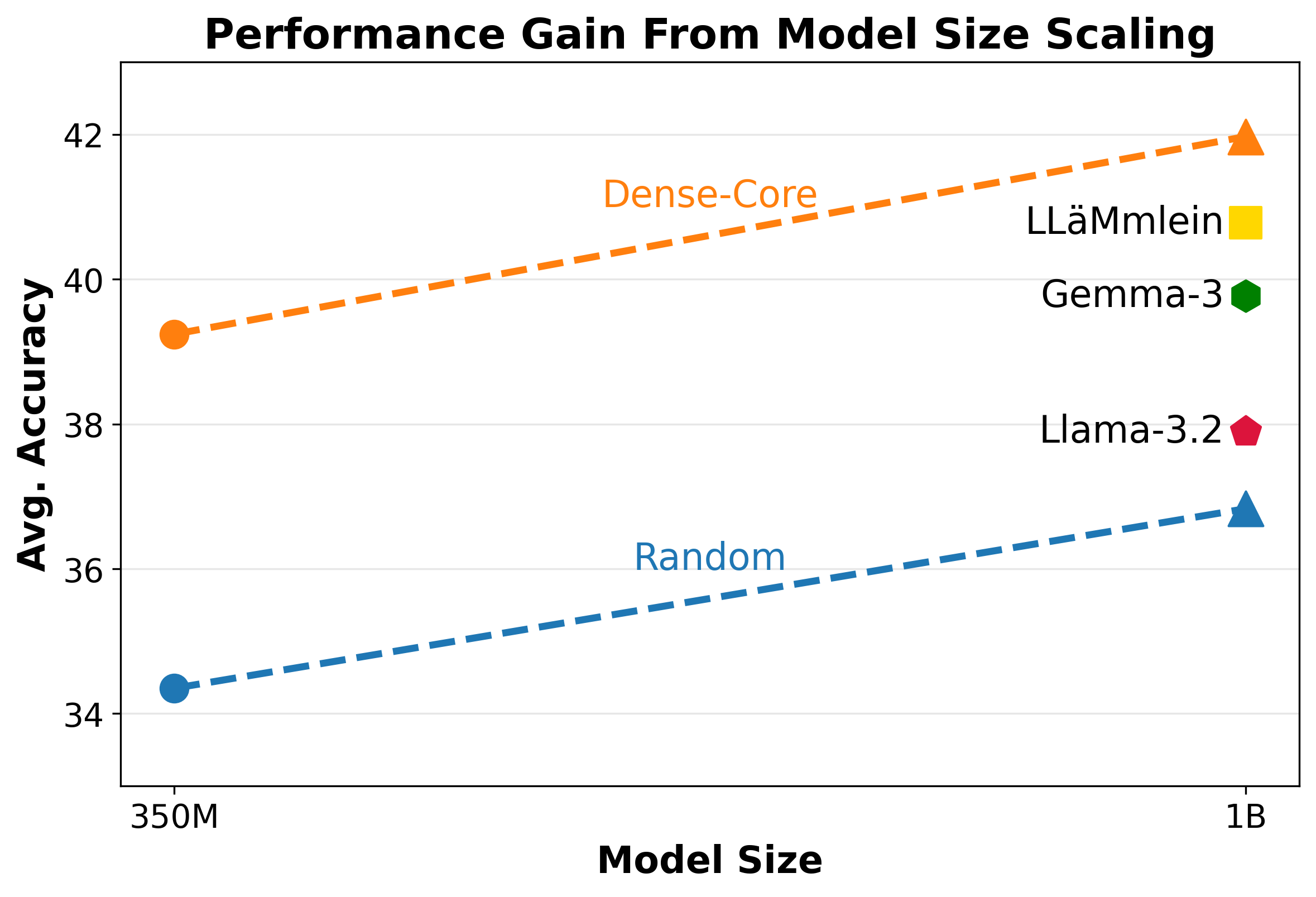}
  \caption{Performance gain when scaling the model size from 350M to 1B parameters. Table \ref{tab:1b-results} breaks down the performance of the models depicted in the graph on each benchmark.}
  \vspace{-4mm}
  \label{fig:results-scaling}
\end{figure}

Figure~\ref{fig:results-scaling} shows that the performance gap between the \textsc{Random} and \textsc{Dense-Core} not just persists, but even slighlty widens as parameter count increases. While the 350M \textsc{Dense-Core} outperformed its baseline by 4.89 points on average, the 1B \textsc{Dense-Core} achieves a 5.14-point lead over the 1B \textsc{Random} baseline.

Furthermore, despite being trained on an order of magnitude less tokens in total, the 1B \textsc{Dense-Core} model outperforms multilingual \textsc{Gemma-3-1B} and \textsc{Llama-3.2-1B}, and monolingual \textsc{LLäMmlein-1B}, all trained on significantly larger token budgets. 

These findings suggest that larger models are more adept at extracting value from high-density sources: a 100B-token budget suffices for state-of-the-art 1B-model performance when the training signal is sufficiently concentrated, indicating that token utility matters more than volume or cross-lingual transfer for non-English high-resource languages like German.

\subsection{Experiment III: Exploring Repetition Limits (200B Budget)}
\label{sec:repetition-ceiling}

While Experiment I established that 3.6 epochs of repetition on high-density data outperforms single pass on diverse data, it remains unclear where the point of diminishing returns lies for various data regimes.
In this experiment, we explore the limits of multi-epoch training on high-density subsets by extending the token budget to 200B, examining whether the benefits of data density would be offset by the lack of novelty, and analyzing the scaling behavior of lower-density sets on the same budget.

\subsubsection{Experimental Setup}
We extend the training of three key configurations from Experiment I to a total of 200B tokens: \textsc{Random}, \textsc{Phased}, and \textsc{Dense-Core}. For the \textsc{Random} baseline this represents a transition from a 100B single-pass to a 200B single-pass (adding new documents into the mix). For the \textsc{Dense-Core}, the 200B budget results in approximately 7.2 training epochs over the 28B-token subset. The \textsc{Phased} curriculum consists of the 100B \textsc{Random} subset folllowed by 100B tokens of continued exposure to the \textsc{Dense-Core} data. We keep the same hyperparameters as in Experiment I to ensure comparable learning dynamics across all runs.

\begin{figure}[t]
  \includegraphics[width=\linewidth]{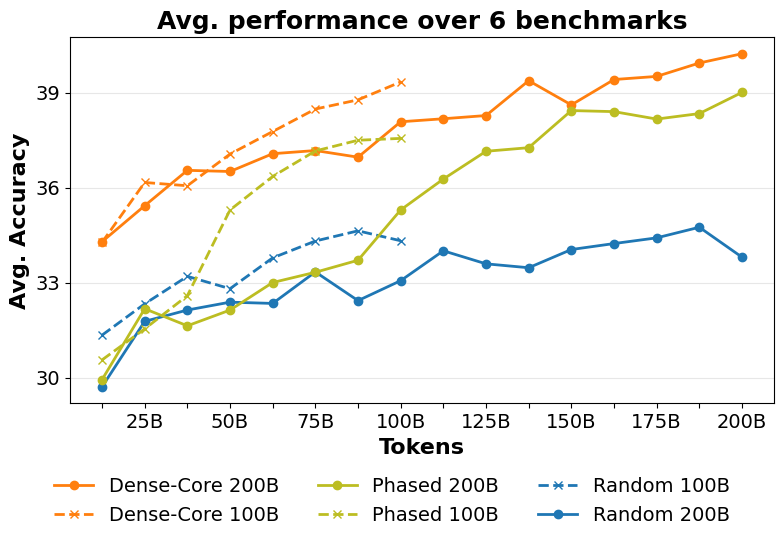}
      \caption{Performance of 350M models trained on 100B and 200B tokens of \textsc{Random} and \textsc{Dense-Core} subsets. Table \ref{tab:200b-results} breaks down the performance of each model's final checkpoint on each benchmark.}
  \vspace{-4mm}
  \label{fig:repetition-ceiling}
\end{figure}


\subsubsection{Results and Analysis}

Figure~\ref{fig:repetition-ceiling} illustrates that the benefits of multi-epoch training on \textsc{Dense-Core} subset persist well beyond the four-epoch threshold identified by \citet{muennighoff-2023-repetition}, extending the findings of \citet{fang2026datasets} to a non-English setting where the unfiltered data pool is severely more constrained.

We do not observe a loss in generalization on our benchmark suite even when doubling the number of epochs on the same dataset. Instead, the \textsc{Dense-Core} model maintains a significant lead over the \textsc{Random} model, even though the latter is still seeing entirely new unique data. While the \textsc{Phased} curriculum continues to scale and narrow the gap to the pure \textsc{Dense-Core} model effectively, it never quite overtakes it, suggesting that training on a more diverse sample in the first stage holds back optimization on a more dense subset.

To investigate whether these trends hold at larger model scales, we conduct an additional pre-training run of a 1B model on 200B tokens of \textsc{Dense-Core}. Table~\ref{tab:1b-results} shows that this model achieves an uplift of 2.08 points on average on our evaluation suite, compared to pre-training on 100B tokens. Notably, this gain is more than twice the corresponding improvement observed for the 350M model.
This indicates that the effectiveness of repeated exposure to high-quality data depends on model size and may extend to longer training horizons as parameter count increases, if the underlying corpus is sufficiently information-dense. In this regime, larger models can continue to extract additional value from repeated passes over curated data, before the requirement for larger pools of unique tokens ultimately becomes the limiting factor.

\subsection{Experiment IV: Generalization to Instruction Tuning}\label{sec:results-it}

Finally, we evaluate whether the advantages of a high-density pre-training translate into improved instruction-following capabilities.

\subsubsection{Experimental Setup} We fine-tune the final checkpoints of our models on the German subset of the \textsc{SmolTalk2} instruction-tuning dataset \cite{bakouch2025smollm3}. It includes 52K German instruction-output pairs and covers a variety of tasks, ranging from creative writing and open-ended QA to coding assistance and math problems. Both the 350M and 1B parameter variants are tuned using identical hyperparameters (see Appendix \ref{app:hparams}). To evaluate the models, we rely on an LLM-as-a-judge protocol that consists of 2 evaluation axes (see Figures \ref{fig:single-prompt} and \ref{fig:binary-prompt} for the prompt templates) and 3 judges: \textsc{Llama-3.3-70B-Instruct}, \textsc{Qwen-3.5-27B}, and \textsc{Gemma-4-31B} \cite{llama3,qwen3.5,gemma4} in order to mitigate potential biases of one specific judge.

Table~\ref{tab:instruction-results} reports the average Likert-scale score (1-10) and share of correct answers (binary accuracy) across 1000 outputs generated from holdout prompts of \textsc{SmolTalk2}.

\begin{table}[t]
\resizebox{1\linewidth}{!}{
\centering
\small
\setlength{\tabcolsep}{4pt} 
\begin{tabular}{lccc}
\toprule
\textbf{Subset} & \textbf{Rep.} & \textbf{Score} & \textbf{Correct} \\
\midrule
\multicolumn{4}{l}{\small \textit{350M params @ 100B tokens}} \\
\textsc{Random} & 1.0x & 3.11 & 9.1 \\
\textsc{Coherence} & 1.0x & 3.47 & 12.4 \\
\textsc{Info. Value} & 1.5x & 3.54 & 13.7 \\
\textsc{Edu. Quality} & 3.2x & \textbf{3.62} & 14.2 \\
\hdashline
\textsc{Dense-Core} & 3.6x & \textbf{3.62} & \textbf{14.4} \\
\textsc{MKC} & 2.9x & 3.36 & 9.4 \\
\textsc{AA High} & 4.8x & 3.24 & 9.6 \\
\hdashline
\textsc{Sorted} & 1.0x & 3.45 & 11.9 \\
\textsc{Phased} & 1.3x & 3.49 & 12.6 \\
\midrule
\multicolumn{4}{l}{\small \textit{350M params @ 200B tokens}} \\
\textsc{Random} & 1.0x & 3.34 & 11.2 \\
\textsc{Dense-Core} & 7.2x & \textbf{3.89} & \textbf{19.7} \\
\textsc{Phased} & 2.6x & 3.67 & 15.0 \\
\midrule
\multicolumn{4}{l}{\small \textit{1B params}} \\
\textsc{Random 100B} & 1.0x & 3.79 & 18.0 \\
\textsc{Dense-Core 100B} & 3.6x & 4.28 & 24.0 \\
\hdashline
\textsc{Dense-Core 200B} & 7.2x & \textbf{4.62} & \textbf{30.7} \\
\midrule
\midrule
\multicolumn{4}{l}{\small \textit{Reference models}} \\
\textsc{Qwen3-0.6B-Base} & - & 4.13 & 18.1 \\
\textsc{LLäMmlein-1B} & - & 4.51 & 28.8 \\
\textsc{Llama-3.2-1B} & - & 4.65 & 27.9 \\
\textsc{Gemma-3-1B} & - & \textbf{4.69} & \textbf{30.8} \\
\bottomrule
\end{tabular}
}
\caption{LLM-as-a-Judge results. \textbf{Rep.} indicates the repetition factor over the unique subset. \textbf{Score} is the 1--10 Likert average; \textbf{Correct} is the average binary accuracy over 1,000 prompts. Refer to Tables \ref{tab:judge_single} and \ref{tab:judge_binary} for per-judge breakdowns of each score.}
\label{tab:instruction-results}
\end{table}

\begin{table*}[ht]
\centering
\resizebox{1\linewidth}{!}{
\begin{tabular}{lccccccccc}
\toprule
\toprule
Model & Tokens & MMLU & ARC-C & ARC-E & H-Swag & LAMBADA & OBQA & Avg. \\
\midrule
\multicolumn{9}{l}{\small \emph{Ours}} \\
\textsc{Boldt-DC-350M} & 200B & 29.29 & 32.24 & 52.87 & 43.21 & 37.48 & 45.86 & 40.16 \\
\textsc{Boldt-DC-1B} & 200B & 31.06 & \textbf{35.99} & \textbf{57.30} & \underline{48.69} & 42.80 & \underline{48.48} & 44.05 \\
\textsc{Boldt-1B} & 230B &\textbf{31.42} & \underline{34.11} & \underline{55.78} & \textbf{48.77} & \underline{44.70} & \textbf{52.32} & \textbf{44.52} \\
\midrule
\multicolumn{9}{l}{\small \emph{Reference models - 1B}} \\
Qwen3.5-0.8B-Base \cite{qwen3.5} & >36T${^*}$ & 30.79 & 32.05 & 46.20 & 38.90 & 36.02 & 43.84 & 37.97 \\
LLäMmlein-1B \cite{llamlein} & 1T & 29.26 & 30.27 & 48.19 & 44.80 & \textbf{44.89} & 47.27 & 40.78 \\
Gemma-3-1B \cite{gemma3} & 2T${^*}$ & 30.01 & 30.55 & 47.89 & 43.43 & \underline{41.71} & 45.05 & 39.77 \\
Llama-3.2-1B \cite{llama3} & 9T${^*}$ & 28.58 & 29.90 & 40.51 & 40.07 & 44.31 & 44.04 & 37.90 \\
\midrule
\midrule
\multicolumn{9}{l}{\small \emph{Reference models - >1B}} \\
EuroLLM-1.7B \cite{eurollm} & 4T${^*}$ & 31.04 & 31.58 & 54.68 & 45.30 & 44.52 & 50.50 & 42.94 \\
Qwen3-1.7B-Base \cite{qwen3} & 36T${^*}$ & 34.17 & 37.49 & 57.00 & 45.20 & 49.81 & 45.66 & 44.89 \\
BübleLM-2B \cite{bueble} & 2T${^*}$ & 29.68 & 32.62 & 53.63 & 46.57 & 43.55 & 49.70 & 42.63 \\
Gemma-2-2B \cite{gemma2} & 2T${^*}$ & 33.99 & 37.11 & 57.47 & 49.62 & 52.64 & 48.89 & 46.62 \\
\bottomrule
\bottomrule
\end{tabular}
}
\caption{Benchmark results of \textsc{Boldt} models compared to other pre-trained models of similar size, as well as larger reference models.}
\vspace{-5mm}
\label{tab:boldt-results}
\end{table*}

\subsubsection{Results and Analysis}
Table~\ref{tab:instruction-results} shows that the advantage of quality filtering for pre-training is preserved through the SFT process. For the 350M models trained on 100B tokens, the \textsc{Dense-Core} configuration achieves the highest average Likert score and share of correct responses, closely followed by \textsc{Educational Quality} and \textsc{Information Value}, significantly exceeding \textsc{Random}.

Interestingly, the \textsc{MKC} subset, which was filtered for documents similar to instruction-tuning datasets like Aya \cite{aya}, fails to compete with our manually defined filters. This suggests that the fundamental reasoning capabilities and factual grounding provided by high-quality, information-rich data are more important for SFT than simply matching the formatting or distribution of typical instruction sets. The poor performance of the \textsc{AA High} subset further reinforces that repetition only benefits sufficiently high-quality data.


\noindent \textbf{Scaling in Parameters and Tokens.} The benefits of semantic density scale effectively across both parameter counts and token budgets. Most strikingly, the 350M \textsc{Dense-Core} model trained for 200B tokens outperforms the 1B \textsc{Random} model trained on 100B tokens both in terms of the Likert Score and share of correct responses despite having $3\times$ fewer parameters.

Unlike the pre-training benchmarks, only models trained on 200B tokens of \textsc{Dense-Core} perform competitively with multilingual reference models after instruction tuning. We attribute this gap to cross-lingual generalization from English data in the reference models' pre-training mixtures, which surfaces during post-training through English-aligned intermediate representations \cite{wendler-etal-2024-llamas, schut2025, park2026collapse}.

\section{Model and Benchmark Release}

We 
release\footnote{\url{https://huggingface.co/Boldt}}
our trained models, called \textsc{Boldt}, together with the filtered FineWeb2 subsets they were trained on, as well as the updated benchmarks from our evaluation suite to the research community. In addition to all the base models we trained as part of our experiments, we release an additional pre-trained model: 

\noindent 
\textbf{\textsc{Boldt-1B}.} A 1B parameter model trained on a combination of \textsc{Dense-Core} and a corpus of 6B tokens of news data in German obtained using the \textsc{Fundus} library~\cite{dallabetta-etal-2024-fundus}. The majority of \textsc{Fundus} news articles stems from the time period between 2022 and February 2026, with smaller numbers of articles going back to 1994. \textsc{Boldt-1B} followed the same multi-epoch approach on the combined corpus until reaching the effective training token count of 230B. The context window size is extended from 2048 to 4096 compared to models trained in Section~\ref{sec:results}.

\noindent
We compare \textsc{Boldt-1B} together with 350M and 1B models trained on the \textsc{Dense-Core} subset for 200B tokens (\textsc{Boldt-DC-350M} \& \textsc{Boldt-DC-1B}) against other similarly-sized pre-trained models on our evaluation suite. The results are shown in Table~\ref{tab:boldt-results}. We find that \textsc{Boldt-1B} improves upon \textsc{Boldt-DC-1B} in all benchmarks except for ARC-Challenge and ARC-Easy. Despite being trained on substantially smaller training corpora compared to other similarly-sized LLMs capable of German, our 1B models are competitive even with larger-sized (\textasciitilde 2B) multilingual models.

\section{Conclusion}

This work addresses a practical question for non-English language modeling: is aggressive quality filtering worthwhile when total available text is limited, or does it discard too much useful diversity for pre-training?

Our experiments show that quality filtering remains beneficial when it comes to constrained data pools of non-English languages like German. Across the model sizes and token budgets we considered, multi-epoch training on small, high-quality subsets consistently outperforms single-pass training on larger, less filtered mixtures, with no early saturation observed. Extending pre-training on the most semantically dense subset we studied yields consistent gains in both pre-training benchmarks and downstream instruction-following, even after 7 epochs of repeated exposure.

In practice, our findings suggest that extensive quality filtering relying on clearly-defined rules and strong annotator models provides a viable path towards sample-efficient pre-training in non-English high-resource languages like German. While this work does not provide direct empirical evidence that our findings generalize to other languages, we see no principled obstacles towards that, as the filtering criteria considered in this work target structural and semantic properties like general text coherence, factual density, amd pedagogical clarity, that are not specific to German. Therefore, the filtering logic is largely language-agnostic, even if the filter design, optimal filtering thresholds or the magnitude of the filtering effect may differ. We leave the validation of our findings across different language families to future work.

\section*{Limitations}

\noindent \textbf{Language scope.} Our investigation focuses exclusively on German as a representative high-resource non-English language. While German’s web corpus characteristics (hundreds of billions of tokens but not trillions) likely generalize to languages like French, Japanese, and Chinese, languages with smaller corpora or different linguistic structures may exhibit different quality-quantity trade-offs.

\noindent Furthermore, we note that our annotation and filtering pipeline relies on multilingual models as judges and classifiers, whose performance is known to vary across languages. The quality of the resulting filtered subsets, and consequently downstream model performance, may therefore differ when applying our pipeline to a language on which the underlying annotator model performs less reliably than on German.

\noindent \textbf{Model and compute scale.} Our experiments are limited to models up to 1B parameters trained on at most 200B tokens. While our findings demonstrate clear efficiency gains at this scale, it remains unclear whether these quality-quantity trade-offs hold at substantially larger, industry-level, scales where both compute budgets and available high-quality data increase dramatically.

\noindent \textbf{Architecture selection.} We focus exclusively on dense transformer architectures. Mixture-of-experts models might exhibit different behaviors with respect to data quality and repetition. Similarly, we do not explore architectural innovations like alternative attention mechanisms that might exhibit different behavior during pre-training.

\noindent \textbf{Toxicity and bias assessment.} We do not evaluate our models for toxic content generation, demographic biases, or harmful stereotypes. While our quality filters prioritize educational and informative content, which may reduce certain risks compared to unfiltered web data, we cannot guarantee that aggressive filtering eliminates problematic content or prevents biased model behavior. 

\noindent High-quality educational content can still encode societal biases, and repeated exposure during multi-epoch training might amplify rather than mitigate such issues. Additionally, our LLM-as-judge evaluation for instruction-tuned models does not assess safety, appropriateness, or potential for harmful outputs. Future work should conduct comprehensive bias audits across demographic dimensions and toxicity benchmarks to understand how quality filtering strategies affect model biases.

\noindent Despite these limitations, our work provides valuable evidence that quality filtering and multi-epoch training offer a practical path toward efficient pre-training for high-resource non-English languages, challenging the notion that maximizing unique token exposure should be the primary objective in data-constrained settings.

\section*{Acknowledgments}

The authors are supported by the Deutsche Forschungsgemeinschaft (DFG, German Research Foundation) under Emmy Noether grant “Eidetic Representations of Natural Language” (project number 448414230). Further, Alan Akbik is supported by the Deutsche Forschungsgemeinschaft (DFG, German Research Foundation) under Germany’s Excellence Strategy "Science of Intelligence" (EXC 2002/1, project number 390523135).

The authors gratefully acknowledge the scientific support and HPC resources provided by the Erlangen National High Performance Computing Center (NHR@FAU) of the Friedrich-Alexander-Universität Erlangen-Nürnberg (FAU) under the NHR project c107fa. NHR funding is provided by federal and Bavarian state authorities. NHR@FAU hardware is partially funded by the German Research Foundation (DFG) – 440719683.

\bibliography{custom}

\clearpage

\appendix

\section{FW2-DE Annotation Details}
\label{app:filter_details}

\begin{figure*}[h]
  \includegraphics[width=\linewidth]{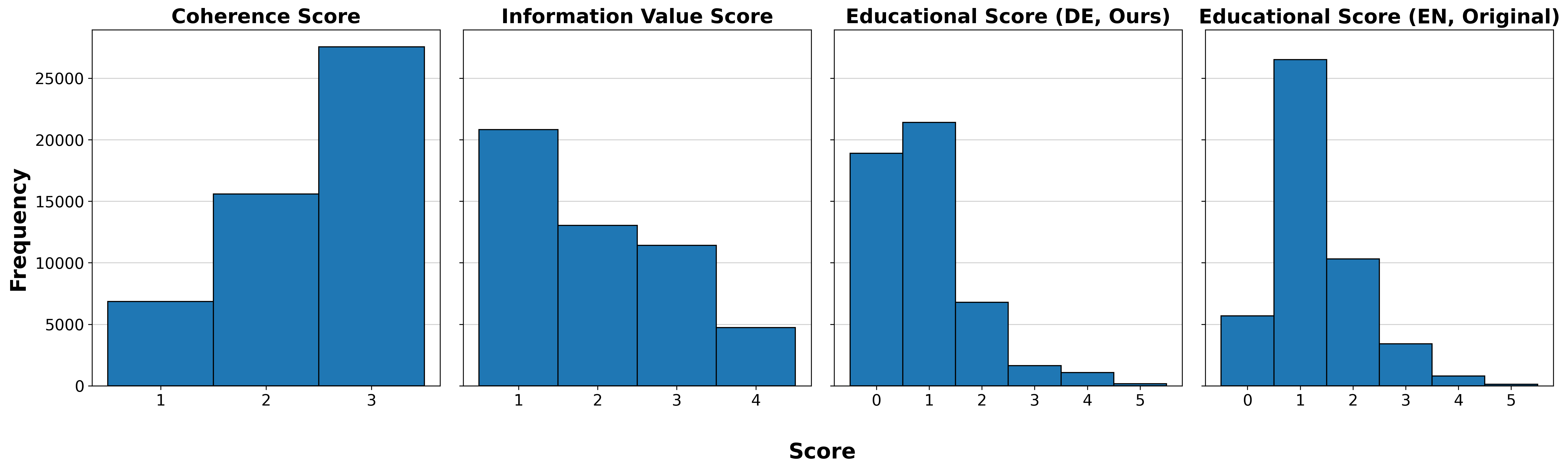}
  \caption{Distribution of Coherence, Information Value, and Educational Quality scores in FW2-DE yielded by Llama-3.3-70B-Instruct annotation. We add the Education Quality score distribution from the original English Fineweb-Edu for reference.}
  \label{fig:annot-label-distr}
\end{figure*}

We closely follow the annotation pipeline outlined by \citet{fineweb-edu} in creation of the English Fineweb-Edu. We first used an LLM to score 500k randomly sampled FW2-DE documents for their Coherence, Information Value, and Educational Quality. For Educational Quality, we use the prompt defined by \citet{fineweb-edu}, while for Coherence and Information Value we define our own prompt, provided in Figure \ref{fig:iv-coh-annotation}. The resulting label distributions are shown in Figure \ref{fig:annot-label-distr}. We truncate input documents to a max length of 4096 tokens.

The LLM that we use for initial annotation is Llama-3.3-70B-Instruct \cite{llama3}. We also considered using Qwen3-32B \cite{qwen3} and Gemma-3-27B \cite{gemma3} due to their extensive multilingual support, however we found their performance to be on average weaker compared to a more capable 70B model, as we show in Table \ref{tab:annotator-eval}, and the resulting distributions to be inferior to the one yielded by the 70B model.

We then fine-tuned three respective \texttt{snowflake-arctic-embed-m-v2.0} \cite{arctic_embed} regression models on the annotated 500k sample, each responsible for scoring one of the document's quality aspects. A hyperparameter sweep showed the linear decay of a max learning rate of 1e-4 to 0 and a batch size of 64 to be optimal over 10 epochs. The best epoch is chosen at the end based on the holdout validation set accuracy. We truncate the web documents to a max length of 1024 tokens. The fine-tuned classifiers were then used annotate the full 500M document German subset of FineWeb2. We report their performance on the holdout test set in Table \ref{tab:classifier_performacne} and release them to the 
public\footnote{\url{https://huggingface.co/Boldt}}.

The \textsc{Coherence} and \textsc{Information Value} subsets are derived by filtering out the documents receiving a score below the respective maximum scores, i.e. 3 and 4, while the \textsc{Educational Quality} subset is derived by applying a threshold of 3. The intersection of these subsets, \textsc{Dense-Core}, is derived by applying all 3 score thresholds at the same time.

\begin{table}[h]
\centering
\small
\begin{tabular}{lcc}
\toprule
\textbf{Classifier} & \textbf{Micro F1} & \textbf{Binary F1} \\
\midrule
\textsc{Coherence} & 84.32 & 90.38 \\
\textsc{Information Value} & 78.45 & 78.50 \\
\textsc{Educational Quality} & 75.61 & 72.32 \\
\bottomrule
\end{tabular}
\caption{Performance of the regression models on the holdout test set. Since the final annotations are discrete score levels, we report both micro F1 (each score level treated as a class, using the rounded predicted score) and binary F1 at each filter's score cutoff used to derive the final subsets (\textsc{Coherence} $\geq 3$, \textsc{Information Value} $\geq 4$, \textsc{Educational Quality} $\geq 3$).}
\label{tab:classifier_performacne}
\end{table}

\section{LLM Training Details}
\label{app:hparams}

We parametrize our 350M models with 24 layers, each with hidden layer size of 1024 and FFN size of 4096, while our 1B models consist of 16 layers, each with hidden layer size of 2048 and FFN size of 8192. All models operate on the context window size of 2048.

In all pre-training runs we use the cosine decay learning rate schedule \cite{loshchilov2017sgdr} with a warmup lasting 1\% of total training steps and decay to 1\% of the peak learning rate, which we empirically determined to be optimal at $5e-4$ for all models. We use the AdamW optimizer \cite{loshchilov2018decoupled} with $\beta_1 = 0.9$, $\beta_2 = 0.95$ and decoupled $L_2$ weight decay coefficient of $0.1$. The effective batch size is 0.5M tokens and the gradients are clipped to a maximal Euclidean norm of $1.0$. 

For instruction tuning we use a similar set of hyperparameters with the main differences in the learning warmup lasting 3\% of training steps, maximum learning rate being equal to $5e-5$, the batch size being 32, and AdamW $\beta_2 = 0.99$. The training loss is calculated only on the instruction completions.

We rely on Huggingface's Nanotron (v0.4), Datatrove (v0.5), and TRL (v0.25.1) libraries to handle the pre-training and instruction-tuning infrastructure. All pre-trianing runs were conducted on 8xA100 and 4xH100 Nvidia GPUs, and all instruction-tuning runs on 1xH100 Nvidia GPU. All models were trained with bfloat16 precision.

For output generation with instruction-tuned variants of our models, we use the same random seed and rely on top-p sampling with $p=0.95$ and temperature $t=0.9$. For LLM-as-a-judge output generation, we force greedy decoding to ensure consistency of evaluation. All open-ended output generation was conducted using the vLLM (v0.19.1) library.

\section{Artifact Licenses}

This work builds on several publicly available datasets and models, all of which are used in accordance with their respective licenses. The primary pre-training corpus, FineWeb2, as well its \textsc{MKC} subset, is released under the ODC-By 1.0 license and is subject to CommonCrawl's Terms of Use. The \textsc{AA High} subset is released under the OpenAlpha 1.0. The instruction-tuning dataset SmolTalk2 is released under Apache 2.0. The \textsc{snowflake-arctic-embed-m-v2.0} model used to train our quality classifiers is released under Apache 2.0. \textsc{Llama-3.3-70B-Instruct}, used for initial data annotation and as one of the LLM-as-a-judge evaluators, is governed by the Llama 3.3 Community License Agreement. The evaluation benchmarks used in this work carry the following licenses: Global MMLU (Apache 2.0), ARC-Easy and ARC-Challenge (CC BY-SA 4.0), HellaSwag (MIT), LAMBADA OpenAI variant (MIT), and OpenBookQA (Apache 2.0). Our re-translated and cleaned versions of the German benchmarks, as well as the model weights we release, will be made available under licenses compatible with their upstream sources.

\begin{table*}[ht]
\resizebox{\linewidth}{!}{
\centering
\begin{tabular}{lcccccc}
\toprule
\textbf{Model} & \textbf{MMLU (DE)} & \textbf{MGSM (DE)} & \textbf{IFEval} & \textbf{Include (DE) } & \textbf{MLQA (DE\_EN)} & Avg. \\
 & (Acc.) & (Acc.) & (Acc.) & (Acc.) & (F1) & \\
\midrule
Gemma-3-27B-IT & 59.09 & 69.60 & 86.21 & 56.18 & 20.70 & 58.36 \\
Qwen3-32B & \textbf{73.59} & 48.40 & 88.49 & 60.93 & 18.79  & 58.04 \\
Llama-3.3-70B-Instruct & 73.13 & \textbf{74.80} & \textbf{93.17} & \textbf{64.07} & \textbf{50.44} & \textbf{71.12} \\
\bottomrule
\end{tabular}
}
\caption{IT-LLM Annotator Evaluation. Model precision: bfloat16. Each benchmark evaluation was done with and without the provided prompt template, and the best result is reported. The Include (DE) results reflect the average over only the STEM and Social Science subsets. IFEval results reflect the "strict" version of the task.}
\label{tab:annotator-eval}
\end{table*}

\clearpage

\begin{table*}[ht]
\centering
\resizebox{1\linewidth}{!}{
\begin{tabular}{lccccccccc}
\toprule
\toprule
Subset & Token Count & MMLU & ARC-Challenge & ARC-Easy & Hellaswag & LAMBADA & OBQA & Avg. \\
\midrule
\multicolumn{9}{l}{\small \emph{Baseline}} \\
\textsc{Random} & 100B (1.0x) & 27.13 & 26.15 & 41.10 & 37.18 & 33.52 & 41.01 & 34.35 \\
\textsc{Random} 197B & 200B (1.0x) & 26.82 & 26.24 & 42.11 & 38.39 & 36.16 & 38.79 & 34.75 \\
\hdashline
\multicolumn{9}{l}{\small \emph{Filter Combinations}} \\
\textsc{Dense-Core} & 28B (3.6x) & 28.97 & 31.40 & 50.55 & 41.10 & \textbf{37.55} & 45.86 & 39.24 \\
\textsc{Dense-Core} 200B & 28B (7.2x) & \textbf{29.29} & \textbf{32.24} & \textbf{52.87} & \textbf{43.21} & 37.48 & 45.86 & \textbf{40.16} \\
\hdashline
\multicolumn{9}{l}{\small \emph{Curriculum-Based}} \\
\textsc{Phased} & 78B (1.3x) & 28.24 & 29.05 & 48.35 & 40.38 & 36.45 & 45.05 & 37.92 \\
\textsc{Phased} 200B & 128B (1.6x) & 28.54 & 30.65 & 51.27 & 40.93 & 36.14 & \textbf{46.87} & 39.07 \\
\bottomrule
\bottomrule
\end{tabular}
}
\caption{Benchmark results of 350M models trained on 200B tokens of different FW2-DE subsets compared to their counterparts trained on 100B tokens of (conceptually) the same subsets. For the model trained on 200B tokens of the \textsc{Random} subset, we report the performance of the 197B checkpoint, as it performs better than the 200B checkpoint on average.}
\label{tab:200b-results}
\end{table*}

\begin{table*}[ht]
\centering
\resizebox{1\linewidth}{!}{
\begin{tabular}{lccccccccc}
\toprule
\toprule
Subset & Token Count & MMLU & ARC-Challenge & ARC-Easy & Hellaswag & LAMBADA & OBQA & Avg. \\
\midrule
\multicolumn{9}{l}{\small \emph{Baseline}} \\
\textsc{Random} 350M & 100B (1.0x) & 27.13 & 26.15 & 41.10 & 37.18 & 33.52 & 41.01 & 34.35 \\
\textsc{Random} 1B & 100B (1.0x) & 27.28 & 27.46 & 43.97 & 41.60 & 38.84 & 41.82 & 36.83 \\
\hdashline
\multicolumn{9}{l}{\small \emph{Filter Combinations}} \\
\textsc{Dense-Core} 350M & 28B (3.6x) & 28.97 & 31.40 & 50.55 & 41.10 & 37.55 & 45.86 & 39.24 \\
\textsc{Dense-Core} 1B & 28B (3.6x) & 29.88 & 32.99 & 54.64 & 45.51 & 40.29 & 48.49 & 41.97 \\
\textsc{Dense-Core} 1B & 28B \textbf{(7.2x)} & \textbf{31.06} & \textbf{35.99} & \textbf{57.30} & \textbf{48.69} & \textbf{42.80} & \textbf{48.48} & \textbf{44.05} \\
\midrule
\midrule
\multicolumn{9}{l}{\small \emph{Reference models}} \\
LLäMmlein-1B & 1T & 29.26 & 30.27 & 48.19 & 44.80 & 44.89 & 47.27 & 40.78 \\
Gemma-3-1B & 2T${^*}$ & 30.01 & 30.55 & 47.89 & 43.43 & 41.71 & 45.05 & 39.77 \\
Llama-3.2-1B & 9T${^*}$ & 28.58 & 29.90 & 40.51 & 40.07 & 44.31 & 44.04 & 37.90 \\
\bottomrule
\bottomrule
\end{tabular}
}
\caption{Benchmark results of 1B models trained on 100B and 200B tokens of the \textsc{Dense-Core} subset compared to their 350M counterparts and similar-sized reference pre-trained models.}
\label{tab:1b-results}
\end{table*}

\onecolumn

\begin{figure}[h]
\centering
\begin{tcolorbox}[colback=white,colframe=black]
Below is an extract from a web page. Evaluate the quality of the content based on its coherence and whether it provides valuable information on any subject matter that a reader may take away after reading it. Valuable information refers to information that is unbiased, non-promotional, and either well-reasoned or well-put. On the other hand, overly promotional materials, personal opinions or experiences without solid foundations, and spam/adult content are not considered valuable information.

\vspace{0.5em}
Assign two scores based on the scoring system described below.

\vspace{0.5em}
Coherence score (range: 1-3)

Indicates whether the extract is logically structured, easy to follow, and clearly articulated.

\vspace{0.5em}
- 1 point: the extract is mostly incoherent. It may contain rambling or non-linear trains of thought; it could include a lot of interruptions or advertisements.

- 2 points: the extract is somewhat coherent, but contains some distracting asides.

- 3 points: the extract is mostly or fully coherent. It consists of complete sentences and logical paragraphs, the ideas are well-put and well-argued, and has minimal to no irrelevant interruptions.

\vspace{0.5em}
Information value score (range: 1-4) 

Measures the amount of unbiased and useful information contained in the extract.

- 1 point: the extract has little to no information value. It contains only promotional information, biased information, or personal views and preferences that are not well argued.

- 2 points: the extract has some information value, but also some clearly biased or promotional information.

- 3 points: the extract has good information value. The content is clearly formulated, well-reasoned, unbiased, and non-promotional.

- 4 points: the extract has exceptional information value. It provides in-depth insights into its subject matter, clearly enhancing a reader's understanding of the topic.

\vspace{0.5em}
The extract: 

\{extract\}

\vspace{0.5em}
After examining the extract:

- Briefly justify your scores, up to 100 words.

- Conclude with the scores using the format: "Coherence score: <Coherence points>. Information value score: <Information value points>"

\end{tcolorbox}
\caption{Annotation prompt used to assign Information Value and Coherence scores.}
\label{fig:iv-coh-annotation}
\end{figure}

\begin{figure}[ht]
\centering
\begin{tcolorbox}[colback=white,colframe=black]
Please act as an impartial judge and evaluate the quality of the response provided by a German AI assistant to the user question displayed below. Your evaluation should consider factors such as the helpfulness, relevance, accuracy, depth, creativity, and level of detail of the response. Begin your evaluation by providing a short explanation. Be as objective as possible. After providing your brief explanation, please rate the response on a scale of 1 to 10 by strictly following this format: "[[rating]]", for example: "Rating: [[5]]". Your evaluation must be performed in English - this includes your explanation, reasoning, and the final verdict. Do not use German at any point.

\vspace{0.5em}
\textbf{[User Question]}\\
\{question\}

\vspace{0.5em}
\textbf{[The Start of Assistant's Answer]}\\
\{answer\}\\
\textbf{[The End of Assistant’s Answer]}

\end{tcolorbox}
\caption{Prompt template used for standalone Likert-scale evaluation of outputs generated by instruction-tuned models. We closely follow the prompt design provided by \citet{zheng2023}.}
\label{fig:single-prompt}
\end{figure}

\begin{figure}[hb]
\centering
\begin{tcolorbox}[colback=white,colframe=black]
Please act as an impartial judge and evaluate the quality of the response provided by a German AI assistant to the user question displayed below. Your evaluation should consider correctness and helpfulness. You will be given a reference answer in addition to the assistant's answer. Your job is to evaluate whether the assistant's answer is correct. Begin your evaluation by comparing the assistant's answer with the reference answer. Identify and correct any mistakes. Do not allow the length of the responses to influence your evaluation. Be as objective as possible. After providing your brief explanation, output your final verdict by strictly following this format: "[[1]]" if the assistant's answer is correct, and "[[0]]" if assistant's answer is incorrect. Your evaluation must be performed in English - this includes your explanation, reasoning, and the final verdict. Do not use German at any point.

\vspace{0.5em}
\textbf{[User Question]}\\
\{question\}

\vspace{0.5em}
\textbf{[The Start of Reference Answer]}\\
\{answer\}\\
\textbf{[The End of Reference Answer]}

\vspace{0.5em}
\textbf{[The Start of Assistant's Answer]}\\
\{answer\}\\
\textbf{[The End of Assistant’s Answer]}

\end{tcolorbox}
\caption{Prompt template used for evaluation of binary correctness of outputs generated by instruction-tuned models.}
\label{fig:binary-prompt}
\end{figure}

\begin{table}[ht]
\centering
\small
\begin{tabular}{lcccc}
\toprule
Subset & Rep. & Llama-3.3-70B & Gemma-4-31B & Qwen3.5-27B \\
\midrule
\multicolumn{5}{l}{\textit{350M params @ 100B tokens}} \\
Random & 1.0x & 5.22 & 2.28 & 1.84 \\
Coherence & 1.0x & 5.60 & 2.64 & 2.18 \\
Info. Value & 1.5x & 5.68 & 2.74 & 2.19 \\
Edu. Quality & 3.2x & \textbf{5.82} & 2.76 & 2.29 \\
\hdashline
Dense-Core & 3.6x & 5.72 & \textbf{2.78} & \textbf{2.36} \\
MKC & 2.9x & 5.64 & 2.48 & 1.96 \\
AA High & 4.8x & 5.36 & 2.41 & 1.95 \\
\hdashline
Sorted & 1.0x & 5.58 & 2.61 & 2.17 \\
Phased & 1.3x & 5.64 & 2.63 & 2.19 \\
\midrule
\multicolumn{5}{l}{\textit{350M params @ 200B tokens}} \\
Random & 1.0x & 5.45 & 2.50 & 2.07 \\
Dense-Core & 7.2x & \textbf{5.92} & \textbf{3.13} & \textbf{2.63} \\
Phased & 2.6x & 5.80 & 2.84 & 2.38 \\
\midrule
\multicolumn{5}{l}{\textit{1B params}} \\
Random 100B & 1.0x & 5.84 & 2.99 & 2.53 \\
Dense-Core 100B & 3.6x & 6.23 & 3.52 & 3.09 \\
\hdashline
Dense-Core 200B & 7.2x & \textbf{6.42} & \textbf{3.93} & \textbf{3.50} \\
\midrule
\multicolumn{5}{l}{\textit{Reference models}} \\
Qwen3-0.6B-Base & -- & 6.80 & 3.14 & 2.44 \\
LLämmlein-1B & -- & 6.33 & 3.80 & 3.39 \\
Llama-3.2-1B & -- & \textbf{6.88} & 3.86 & 3.21 \\
Gemma-3-1B & -- & 6.71 & \textbf{3.94} & \textbf{3.41} \\
\bottomrule
\end{tabular}
\caption{LLM-as-a-judge average Likert scores (1--10, higher is better).}
\label{tab:judge_single}
\end{table}

\begin{table}[ht]
\centering
\small
\begin{tabular}{lcccc}
\toprule
Subset & Rep. & Llama-3.3-70B & Gemma-4-31B & Qwen3.5-27B \\
\midrule
\multicolumn{5}{l}{\small \textit{350M params @ 100B tokens}} \\
Random & 1.0x & 100 & 123 & 50 \\
Coherence & 1.0x & 134 & 166 & 71 \\
Info. Value & 1.5x & 146 & 184 & 82 \\
Edu. Quality & 3.2x & \textbf{161} & 177 & 88 \\
\hdashline
Dense-Core & 3.6x & 142 & \textbf{195} & \textbf{95} \\
MKC & 2.9x & 98 & 140 & 43 \\
AA High & 4.8x & 95 & 136 & 58 \\
\hdashline
Sorted & 1.0x & 138 & 155 & 65 \\
Phased & 1.3x & 130 & 172 & 77 \\
\midrule
\multicolumn{5}{l}{\small \textit{350M params @ 200B tokens}} \\
Random & 1.0x & 110 & 160 & 65 \\
Dense-Core & 7.2x & \textbf{213} & \textbf{260} & \textbf{117} \\
Phased & 2.6x & 177 & 184 & 88 \\
\midrule
\multicolumn{5}{l}{\small \textit{1B params}} \\
Random 100B & 1.0x & 177 & 238 & 126 \\
Dense-Core 100B & 3.6x & 256 & 293 & 171 \\
\hdashline
Dense-Core 200B & 7.2x & \textbf{319} & \textbf{368} & \textbf{235} \\
\midrule
\multicolumn{5}{l}{\small \textit{Reference models}} \\
Qwen3-0.6B-Base & -- & 242 & 209 & 91 \\
LLämmlein-1B & -- & 304 & 347 & 214 \\
Llama-3.2-1B & -- & \textbf{349} & 329 & 159 \\
Gemma-3-1B & -- & 343 & \textbf{366} & \textbf{215} \\
\bottomrule
\end{tabular}
\caption{LLM-as-a-judge binary accuracy (number of correct outputs out of 1000).}
\label{tab:judge_binary}
\end{table}

\end{document}